\documentclass[runningheads]{llncs}

\usepackage{times}
\usepackage{epsfig}
\usepackage{graphicx}
\usepackage{amsmath}
\usepackage{amssymb}
\usepackage{booktabs} 
\usepackage{algorithm}
\usepackage{algorithmic}
\usepackage{subfigure}
\usepackage{url}
\usepackage{multirow}
\usepackage[usenames,dvipsnames,table]{xcolor}

\usepackage{dsfont}
\usepackage{amsmath,amssymb} 
\usepackage{color}
\usepackage[width=122mm,left=12mm,paperwidth=146mm,height=193mm,top=12mm,paperheight=217mm]{geometry}

\begin{document}
\pagestyle{headings}
\mainmatter
\def\ECCV16SubNumber{305}  

\title{Semantic Object Parsing with Graph LSTM} 

\titlerunning{Semantic Object Parsing with Graph LSTM}

\authorrunning{Semantic Object Parsing with Graph LSTM}

\author{Xiaodan~Liang$^{\dagger}$ $^{\star}$ \quad Xiaohui Shen$^{\ast}$  \quad Jiashi Feng$^{\dagger}$ \quad Liang~Lin$^{\star}$ \quad Shuicheng~Yan $^{\dagger}$}
\institute{
		$^{\dagger}$ National University of Singapore \quad $^{\star}$ Sun Yat-sen University \quad $^{\ast}$ Adobe Research\\
	}

\maketitle

\begin{abstract}

By taking the semantic object parsing task as an exemplar application scenario, we propose the Graph Long Short-Term Memory (Graph LSTM) network, which is the generalization of LSTM from sequential data or multi-dimensional data to general graph-structured data. Particularly, instead of evenly and fixedly dividing an image to pixels or patches in existing multi-dimensional LSTM structures (e.g., Row, Grid and Diagonal LSTMs~\cite{gridlstm}\cite{van2016pixel}), we take each arbitrary-shaped superpixel as a semantically consistent node, and adaptively construct an undirected graph for each image, where the spatial relations of the superpixels are naturally used as edges. Constructed on such an adaptive graph topology, the Graph LSTM is more naturally aligned with the visual patterns in the image (e.g., object boundaries or appearance similarities) and provides a more economical information propagation route. Furthermore, for each optimization step over Graph LSTM, we propose to use a confidence-driven scheme to update the hidden and memory states of nodes progressively till all nodes are updated. In addition, for each node, the forgets gates are adaptively learned to capture different degrees of semantic correlation with neighboring nodes. Comprehensive evaluations on four diverse semantic object parsing datasets well demonstrate the significant superiority of our Graph LSTM over other state-of-the-art solutions.

\keywords{Object Parsing, Graph LSTM, Recurrent Neural Networks}
\end{abstract}

\section{Introduction}

\begin{figure}[!tp]
	\begin{center}
		\includegraphics[scale=0.48]{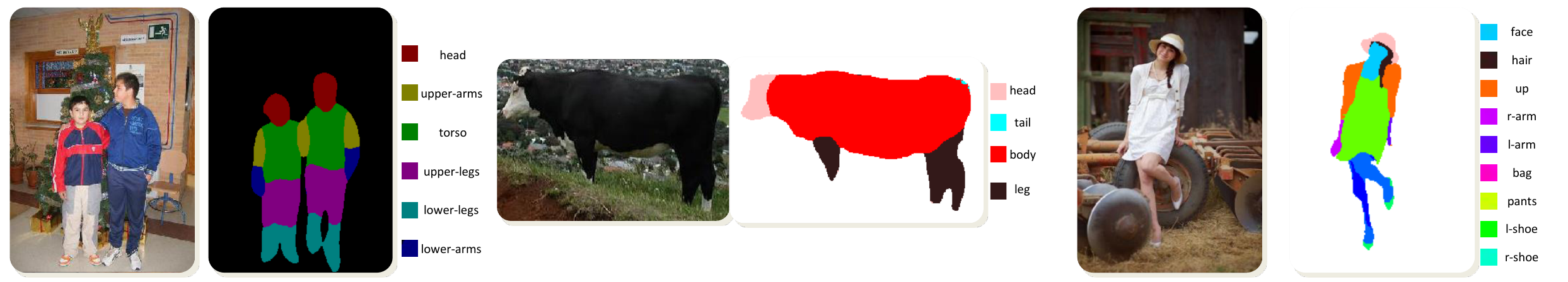}
		\caption{{Examples of semantic object parsing results by the proposed Graph LSTM model. It parses an object into multiple parts with different semantic meanings. Best viewed in color.}}
		\label{fig:task}
	\end{center}
	\vspace{-6mm}
\end{figure}

Beyond traditional image semantic segmentation, semantic object parsing aims to segment an object within an image into multiple parts with more fine-grained semantics and provide full understanding of image contents, as shown in Fig.~\ref{fig:task}. Many higher-level computer vision applications can benefit from a powerful semantic object parser, including action recognition~\cite{sharma2015action}, clothes recognition and retrieval~\cite{Yamaguchiparsing13} and human behavior analysis~\cite{wang2012discriminative}.

Recently, Convolutional Neural Networks (CNNs) have demonstrated exciting success in various
pixel-wise prediction tasks such as semantic segmentation~\cite{crfasrnn}\cite{liu2015semantic}, semantic part segmentation~\cite{xia2015zoom}\cite{chen2015attention} and depth prediction~\cite{eigen2014predicting}. However, the pure convolutional filters can only capture limited local context while the precise inference for semantic part layouts and their interactions requires a global perspective of the image. For example, distinguishing ``upper-arms" from ``lower-arms"  or ``upper-legs" in object parsing needs the sensing of relative spatial layouts and the guidance from the predictions of other semantic regions such as ``torso". To consider the  global structural context, previous works thus use dense pairwise connections (Conditional Random Fields (CRFs)) upon pure pixel-wise CNN classifiers~\cite{liu2015semantic}\cite{chen2014semantic}\cite{crfasrnn}\cite{wang2015joint}\cite{gadde2015superpixel}. However, most of them try to model the structure information based on the predicted confidence maps, and do not explicitly enhance the feature representations in capturing global contextual information, leading to suboptimal segmentation results under complex scenarios. 

\begin{figure*}[!tp]
	\begin{center}
		\includegraphics[scale=0.65]{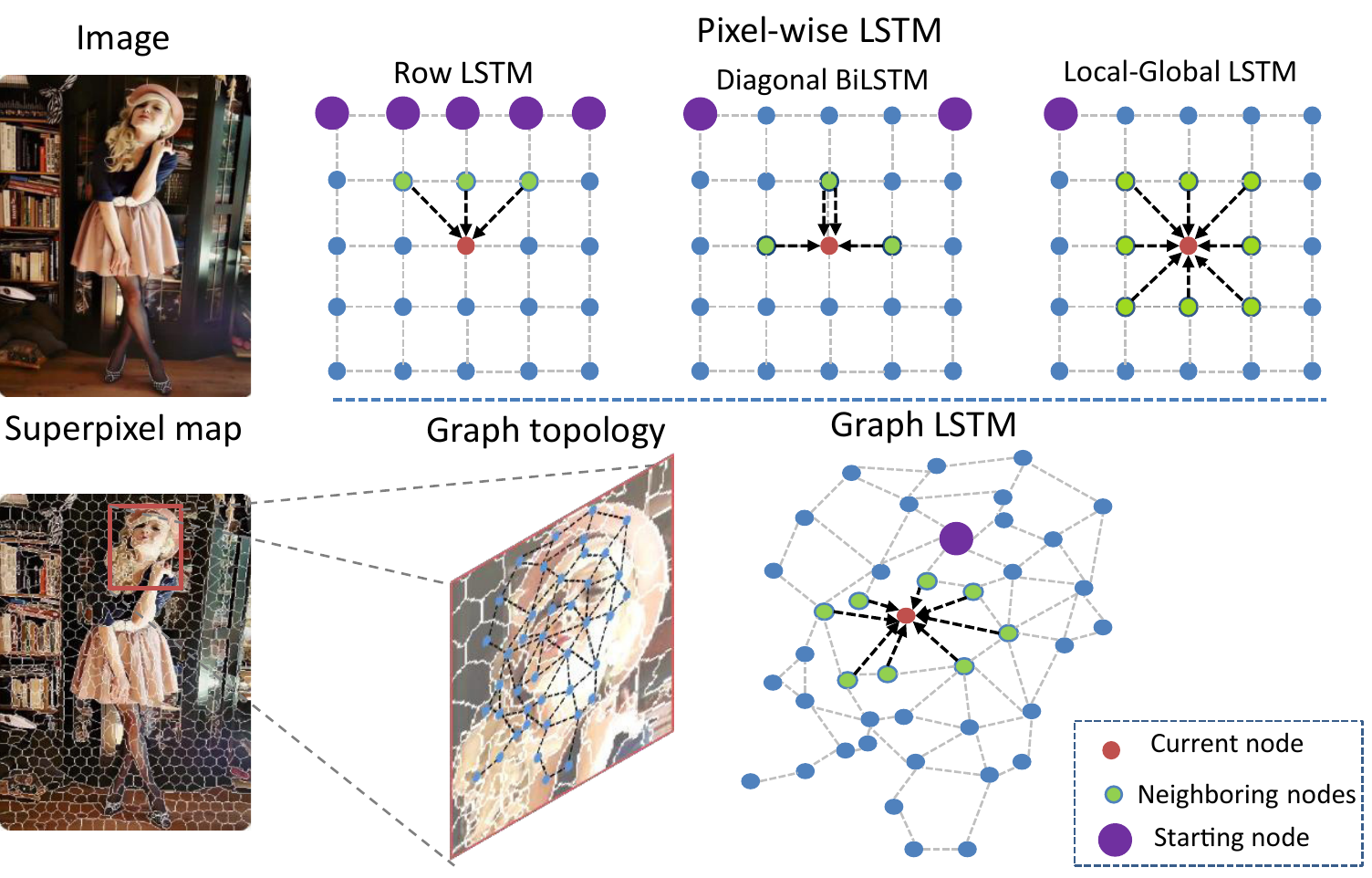}
		\vspace{-3mm}
		\caption{{The proposed Graph LSTM structure. 1) The top row shows the traditional pixel-wise LSTM structures that use the fixed locally factorizations to update the states of each pixel, including Row LSTM~\cite{van2016pixel}, Diagonal BiLSTM~\cite{van2016pixel}~\cite{gridlstm} and Local-Global LSTM~\cite{liang2015semantic}. 2) The bottom row illustrates the proposed Graph LSTM that is built upon the superpixel over-segmentation map for each image. The adaptive graph topology is constructed based on the superpixel nodes and their spatial connections. Each node can thus be influenced by various numbers of neighboring nodes. Instead of using the fixed starting nodes and updating route for all images, the starting node and node updating scheme of Graph LSTM is dynamically specified for each image. }}
		\label{fig:graphlstm}
	\end{center}
	\vspace{-4mm}
\end{figure*}

An alternative strategy is to exploit long-range dependencies by directly augmenting the intermediate features. The multi-dimensional Long Short-Term Memory (LSTM) networks have produced very promising results in modeling 2D images~\cite{byeon2014texture}\cite{theis2015generative}\cite{byeon2015scene}\cite{liang2015semantic}, where long-range
dependencies, which are essential to object and scene understanding, can be well memorized by sequentially functioning on all pixels. However, in terms of the information propagation route in each LSTM unit, most of existing LSTMs~\cite{gridlstm}~\cite{van2016pixel}~\cite{liang2015semantic} have only explored  pre-defined fixed topologies. As illusrated in the top row of Fig.~\ref{fig:graphlstm}, for each individual image, the prediction for each pixel by those methods is influenced by the predictions of fixed neighbors (e.g., 2 or 8 adjacent pixels or diagonal neighbors) in each time-step. The natural properties of images (e.g., local boundaries and semantically consistent groups of pixels) have not be fully utilized to enable more meaningful and economical inference in such fixed locally factorized LSTMs. In addition, much computation with the fixed topology is redundant and  inefficient as it has  to consider  all the pixels, even for the ones in a simple plain region.   

In this paper, we propose a novel Graph LSTM model that extends the traditional LSTMs from sequential and multi-dimensional data to general graph-structured data, and demonstrate its superiority on the semantic object parsing task. Instead of evenly and fixedly dividing an image into pixels or patches as previous LSTMs did, Graph LSTM takes each arbitrary-shaped superpixel as a semantically consistent node of a graph, while the spatial neighborhood relations are naturally used to construct the undirected graph edges. The adaptive graph topology can thus be constructed where different nodes are connected with different numbers of neighbors, depending on the local structures in the image. As shown in the bottom row of Fig.~\ref{fig:graphlstm}, instead of broadcasting information to a fixed local neighborhood following a fixed updating sequence as in the previous LSTMs, Graph LSTM proposes to effectively propagate information from one adaptive starting superpixel node to all superpixel nodes along the adaptive graph topology for each image. It can effectively reduce redundant computational costs while better preserving object/part boundaries to facilitate global reasoning over the whole image. 

Together with the adaptively constructed graph topology of an image, we propose a confidence-driven scheme to subsequently update the features of all nodes, which is inspired by the recent visual attention models~\cite{sharma2015action}\cite{mnih2014recurrent}. Previous LSTMs~\cite{gridlstm}\cite{van2016pixel} often simply start at pre-defined pixel or patch locations and then proceed toward other pixels or patches following a fixed updating route for different images. In contrast, we assume that starting from a proper superpixel node and updating the nodes following a certain content-adaptive path can lead to a more flexible and reliable inference for global context modelling, where the visual characteristics of each image can be better captured. Specifically, for each image, the superpixel node that has the highest predicted confidence across all the foreground semantic labels based on the initial features is regarded as the starting node. Meanwhile, the order of node updating is determined by ranking all the nodes according to their initial confidences on foreground classes in a descending order.


In traditional LSTMs~\cite{gridlstm}\cite{van2016pixel}, by sharing the forget gates, each node receives influence equally from all of its neighboring nodes, which is not always true in visual applications. For example, given a superpixel node in a semantic region, other neighboring superpixels belonging to the same semantic region could provide stronger cues for the  local prediction of the current node than neighboring background superpixel nodes do. Therefore, in Graph LSTM, we adaptively learn the forget gates with respect to different neighboring nodes when updating the hidden states of a certain node in order to model various neighbor connections. Such an adaptive scheme is especially beneficial with Graph LSTM, in which the connections between nodes convey more semantically meaningful interactions than the ones in the pixels/patches based LSTMs with a fixed topology. 

\begin{figure*}[!tp]
	\begin{center}
		\includegraphics[scale=0.55]{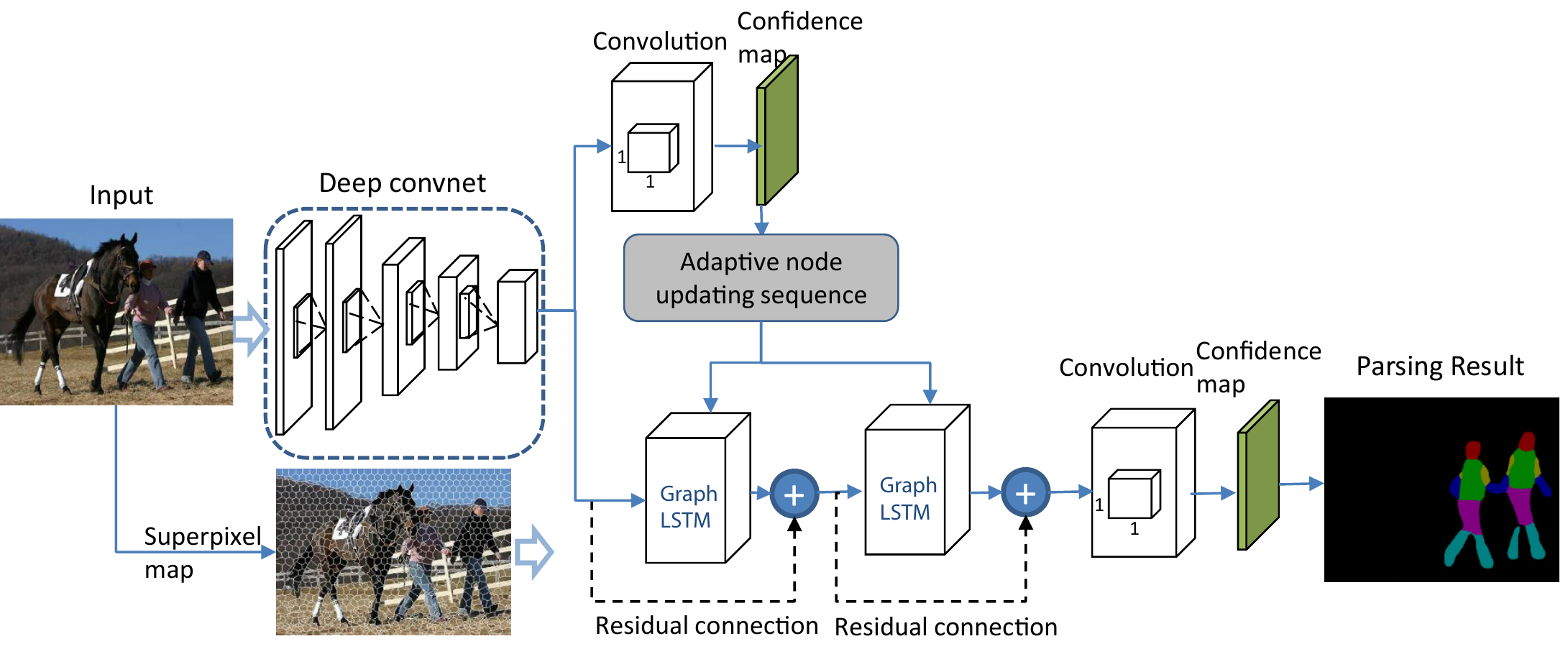}
		\vspace{-2mm}
		\caption{{Illustration of the proposed network architecture for semantic object parsing. The Graph LSTM layers built on a superpixel map are appended on the convolutional layers to enhance visual features with global structure context. The convolutional features pass through $1\times 1$ convolutional filters to generate the initial confidence maps for all labels. The node updating sequence for the subsequent Graph LSTM layers is determined by the confidence-drive scheme based on the initial confidence maps, and then the Graph LSTM layers can sequentially update the hidden states of all superpixel nodes. The residual connections are also incorporated between the Graph LSTM layers to improve the network training with many layers. }}
		\label{fig:architecture}
	\end{center}
	\vspace{-6mm}
\end{figure*}

As shown in Fig.~\ref{fig:architecture}, the Graph LSTM, as an independent layer, can be easily appended to the intermediate convolutional layers in a Fully Convolutional Neural Network~\cite{long2014fully} to strengthen visual feature learning by incorporating long-range contextual information. The hidden states represent the reinforced features, and the memory states recurrently encode the global structures.

Our  contributions can be summarized in the following four aspects. 1)  We propose a novel Graph LSTM structure to extend the traditional LSTMs from sequential and multi-dimensional data to general graph-structured data, which effectively exploits global context by following an adaptive graph topology derived from the content of each image. 2) We propose a confidence-driven scheme to select the starting node and sequentially update all nodes, which facilitates the flexible inference while preserving the visual characteristics of each image. 3) In each Graph LSTM unit, different forget gates for the neighboring nodes are learned to dynamically incorporate the local contextual interactions in accordance with their semantic relations. 4) We apply the proposed Graph LSTM in semantic object parsing, and demonstrate its superiority through comprehensive comparisons on four challenging semantic object parsing datasets (i.e., PASCAL-Person-Part dataset~\cite{chen2014detect}, Horse-Cow parsing dataset~\cite{wang2014semantic}, ATR dataset~\cite{ATR} and Fashionista dataset~\cite{yamaguchi2012parsing}).

\section{Related Work}

\textbf{LSTM on Image Processing:} Recurrent neural networks have been first introduced to address the sequential prediction tasks~\cite{graves2009offline}~\cite{sutskever2014sequence}~\cite{showtell}, and then extended to multi-dimensional image processing tasks~\cite{byeon2014texture}~\cite{theis2015generative} such as image generation~\cite{van2016pixel}\cite{theis2015generative}, person detection~\cite{stewart2015end}, scene labeling~\cite{byeon2015scene} and object parsing~\cite{liang2015semantic}. Benefiting from the long-range memorization of LSTM networks, they can obtain considerably larger dependency fields by sequentially performing LSTM units on all pixels, compared to the local convolutional filters. Nevertheless, in each LSTM unit, the prediction of each pixel is affected by a fixed factorization (e.g., 2 or 8 neighboring pixels~\cite{gridlstm}\cite{MDLSTM}\cite{liang2015semantic} or diagonal neighborhood~\cite{van2016pixel}\cite{theis2015generative}), where diverse natural visual correlations (e.g., local boundaries and homogeneous regions) have not been considered. Meanwhile, the computation is very costly and redundant due to the sequential computation on all pixels.  Different from using locally fixed factorized LSTM units, we propose a novel Graph LSTM structure, which performs the information propagation on varying graph topologies on the superpixel nodes with compact representation. Tree-LSTM~\cite{tai2015improved} introduces the structure with tree-structured topologies for predicting semantic representations of sentences. Compared to Tree-LSTM, Graph LSTM is more natural and general for 2D image processing with arbitrary graph topologies and adaptive updating schemes.

\noindent\textbf{Semantic Object Parsing:} There has been increasing research interest on the semantic object parsing problem including the general object parsing~\cite{wang2014semantic}\cite{wang2015joint}\cite{lu2014parsing}\cite{chen2014detect}\cite{hariharan2014hypercolumns}, person part segmentation~\cite{chen2015attention}\cite{xia2015zoom} and human parsing~\cite{yamaguchi2012parsing}\cite{Yamaguchiparsing13}\cite{Dongparsing13}\cite{ICCV11WBW}\cite{SimoSerraACCV2014}\cite{M-CNN}\cite{Co-CNN}. To capture the rich structure information based on the advanced CNN architecture, one common way is the combination of CNNs and CRFs~\cite{chen2014semantic}\cite{crfasrnn}\cite{schwing2015fully}\cite{wang2015joint}, where the CNN outputs are treated as unary potentials while CRF further incorporates pairwise or higher order factors. Instead of learning features only from local convolutional
kernels as in these previous methods, we incorporate the global context by the novel Graph LSTM structure to capture long-distance dependencies on the superpixels. The dependency field of Graph LSTM can effectively cover the entire image context.

\vspace{-3mm}
\section{The Proposed Graph LSTM}
\vspace{-3mm}

In introducing Graph LSTM, we take semantic object parsing as its application scenario, which aims to generate pixel-wise semantic part segmentation for each image. Fig.~\ref{fig:architecture} illustrates the designed network architecture based on Graph LSTM.  The input image first passes through a stack of convolutional layers to generate the convolutional feature maps. Then the proposed Graph LSTM layers are incorporated to exploit global structure context upon the convolutional feature maps for better fine-grained prediction, building on a generated superpixel map. The proposed Graph LSTM takes the convolutional features and the adaptively specified node updating sequence for each image as the input, and then efficiently propagates the aggregated contextual information towards all nodes, leading to enhanced visual features and better parsing results. To both increase convergence speed and propagate signals more directly through the network, we deploy residual connections~\cite{he2015deep} after one Graph LSTM layer to generate the input features of the next Graph LSTM layer. Note that residual connections are performed to generate the element-wise input features for each layer, which would not destroy the computed graph topology. After that, several $1\times1$ convolution filters are employed to produce the final parsing results. The following subsections will describe the main innovations inside Graph LSTM, including the graph construction and the Graph LSTM structure. 

\subsection{Graph Construction}

The graph is constructed on superpixels that are obtained through image over-segmentation using SLIC~\cite{achanta2010slic}\footnote{Other over-segmentation methods such as entropy rate-based approach [41] could also be used, and we did not observe much difference in the final results in our experiments.}. Note that, after several convolutional layers, the feature maps of each image have been down-sampled. Therefore, in order to use the superpixel map for graph construction in each Graph LSTM layer, one needs to upsample the feature maps into the original size of the input image.


The superpixel graph $\mathcal{G}$ for each image is then constructed by connecting a set of  graph nodes $\{v_i\}_{i=1}^N$ via the graph edges $\{\mathcal{E}_{ij}\}$. Each graph node $v_i$ represents a  superpixel and each graph edge $\mathcal{E}_{ij}$ only connects two spatially neighboring superpixel nodes.  The input features of each graph node $v_i$ are denoted as $\mathbf{f}_i \in \mathbb{R}^d$, where $d$ is the feature dimension. The feature $\mathbf{f}_i$ is computed by averaging the features of all the pixels belonging to the same superpixel node $v_i$. As shown in Fig.~\ref{fig:architecture}, the input states of the first Graph LSTM layer come from the previous convolutional feature maps. For the subsequent Graph LSTM layers, the input states are generated after the residual connections~\cite{he2015deep} for the input features and the updated hidden states by the previous Graph LSTM layer. To make sure that the  number of the input states for the first Graph LSTM layer is compatible with that of the following layers and  that the residual connections can be applied, the dimensions of hidden and memory states in all Graph LSTM layers are set the same as the feature dimension of the last convolutional layer before the first Graph LSTM layer. 

\subsection{Graph LSTM}


\textbf{Confidence-driven Scheme.} The node updating scheme is more important yet more challenging in Graph LSTM than the ones in traditional LSTMs~\cite{van2016pixel}\cite{gridlstm} due to its adaptive graph topology. To enable better global reasoning, Graph LSTM specifies the adaptive starting node and node updating sequence for the information propagation of each image. Given the constructed undirected graph $\mathcal{G}$, we extensively tried several schemes to update all nodes in a graph in the experiments, including the Breadth-First Search (BFS), Depth-First Search (DFS) and Confidence-Driven Search (CDS). We find that the CDS achieves better performance. Specifically, as illustrated in Fig.~\ref{fig:architecture}, given the top convolutional feature maps, the $1\times 1$ convolutional filters can be used to generate the initial confidence maps with regard to each semantic label. Then the confidence of each superpixel for each label is computed by averaging the confidences of its contained pixels, and the label with highest confidence could be assigned to the superpixel. Among all the foreground superpixels (i.e., assigned to any semantic part label), the node updating sequence can be determined by ranking all the superpixel nodes according to the confidences of their assigned labels. The ones with higher confidences will be updated first. If two nodes have the same confidence score, the spatially left one will be first updated. The CDS scheme can provide a relatively more reliable updating sequence for better semantic reasoning, since the earlier nodes in the updated sequence presumably have stronger semantic evidence (e.g., belonging to any important semantic parts with higher confidence) and their visual features may be more reliable for message passing. 

During updating, the $(t+1)$-th Graph LSTM layer determines the current states of each node $v_i$ that comprises the hidden states $\mathbf{h}_{i,t+1} \in \mathbb{R}^d$ and memory states $\mathbf{m}_{i,t+1}\in \mathbb{R}^d$ of each node. Each node is influenced by its previous states and the states of neighboring graph nodes as well in order to propagate information to the whole image. Thus the inputs to  Graph LSTM units consist of the input states $\mathbf{f}_{i,t+1}$ of the node $v_i$,  its previous hidden states $\mathbf{h}_{i,t}$ and memory states $\mathbf{m}_{i,t}$, and the hidden and memory states of its neighboring nodes $v_j, j \in \mathcal{N}_{\mathcal{G}}(i)$.

\textbf{Averaged Hidden States for Neighboring Nodes.} Note that with an adaptive updating scheme, when operating on a specific node in each Graph LSTM layer, some of its neighboring nodes have already been updated while others may have not. We therefore use a visit flag $q_j$ to indicate whether the graph node $v_j$ has been updated, where $q_j$ is set as 1 if updated, and otherwise 0. We then use the updated hidden states $\mathbf{h}_{j,t+1}$ for the visited nodes, i.e., $q_j = 1$ and the previous states $\mathbf{h}_{j,t}$ for the unvisited nodes. The $\mathds{1}(\cdot)$ is an indicator function. Note that the nodes in the graph may have an arbitrary number of neighboring nodes. Let $|\mathcal{N}_{\mathcal{G}}(i)|$ denote the number of neighboring graph nodes. To obtain a fixed feature dimension for the inputs of the Graph LSTM unit during network training, the hidden states $\bar{\mathbf{h}}_{i,t}$ used for computing the LSTM gates of the node $v_i$ are obtained by averaging the hidden states of neighboring nodes, computed as:

\vspace{-4mm}
{\small
\begin{equation}
    \bar{\mathbf{h}}_{i,t} = \frac{\sum_{j \in \mathcal{N}_{\mathcal{G}}(i) }(\mathds{1}(q_j = 1)\mathbf{h}_{j,t+1} + \mathds{1}(q_j = 0)\mathbf{h}_{j,t})}{|\mathcal{N}_{\mathcal{G}}(i)|}.
\end{equation}
}\vspace{-4mm}

\textbf{Adaptive Forget Gates.} Note that unlike the traditional LSTMs~\cite{gridlstm}\cite{lstm}, the Graph LSTM specifies different forget gates for different neighboring nodes by functioning the input states of the current node with their hidden states, defined as $\bar{g}^f_{ij}, j \in \mathcal{N}_{\mathcal{G}}(i)$. It results in the different influences of neighboring nodes on the updated memory states $\mathbf{m}_{i, t+1}$ and hidden states $\mathbf{h}_{i, t+1}$. The memory states of each neighboring node are also utilized to update the memory states $\mathbf{m}_{i, t+1}$ of the current node. The shared weight metrics $U^{fn}$ for all nodes are learned to guarantee the spatial transformation invariance and enable the learning with various neighbors. The intuition is that each pair of neighboring superpixels may be endowed with distinguished semantic correlations compared to other pairs. For instance, two superpixels of the same semantic part should have consistent predictions while two superpixels of different parts can provide contextual cues. The Graph LSTM thus incorporates these adaptive forget gates to cover diverse visual patterns.

\textbf{Graph LSTM Unit.} The Graph LSTM consists of four gates:  the input gate $g^u$, the forget gate $g^f$, the adaptive forget gate $\bar{g}^f$, the memory gate $g^c$ and the output gate $g^o$. The $W^u, W^f, W^c, W^o$ are the recurrent gate weight matrices specified for input features while $U^u,U^f, U^c, U^o$ are those for hidden states of each node. $U^{un}, U^{fn}, U^{cn}, U^{on}$ are the weight parameters specified for states of neighboring nodes. The hidden and memory states by the Graph LSTM can be updated as follows:
{\small\begin{equation}
\begin{split}
g^u_i = &\delta(W^u\mathbf{f}_{i,t+1} + U^u\mathbf{h}_{i,t} + U^{un}\bar{\mathbf{h}}_{i,t} + b^u),\\
\bar{g}^f_{ij} =& \delta(W^f\mathbf{f}_{i,t+1} + U^{fn}\mathbf{h}_{j,t} + b^f),\\
g^f_{i} = &\delta(W^f\mathbf{f}_{i,t+1} + U^f\mathbf{h}_{i,t} + b^f),\\
g^o_i = &\delta(W^o\mathbf{f}_{i,t+1} + U^o\mathbf{h}_{i,t} + U^{on}\bar{\mathbf{h}}_{i,t} + b^o),\\
g^c_i = &\tanh(W^c\mathbf{f}_{i,t+1}  + U^c\mathbf{h}_{i,t} + U^{cn}\bar{\mathbf{h}}_{i,t} + b^c),\\
\mathbf{m}_{i, t+1} = &\frac{\sum_{j \in \mathcal{N}_{\mathcal{G}}(i)}(\mathds{1}(q_j = 1) \bar{g}^f_{ij} \odot \mathbf{\mathbf{m}}_{j,t+1} + \mathds{1}(q_j = 0)\bar{g}^f_{ij} \odot \mathbf{\mathbf{m}}_{j,t})}{|\mathcal{N}_{\mathcal{G}}(i)|}\\ 
& +  g^f_{i}\odot \mathbf{\mathbf{m}}_{i,t} +  g^u_i \odot g^c_i,\\
\mathbf{h}_{i, t+1} =& \tanh(g^o_i \odot \mathbf{m}_{i,t+1}).
\end{split}
\label{eq:lstm}
\end{equation}}
\vspace{-3mm}

\noindent{Here} $\delta$ is the logistic sigmoid function, and $\odot$ indicates a point-wise product. The memory states $\mathbf{m}_{i, t+1}$ of the node $v_i$ are updated by combining the memory states of visited nodes and those of unvisited nodes by using the adaptive forget gates. Let $\mathbf{W}, \mathbf{U}$ denote the concatenation of all weight matrices and $\{\mathbf{Z}_{j,t}\}_{j \in \mathcal{N}_{\mathcal{G}}(i)}$ represent all related information of neighboring nodes. We can thus use $\text{G-LSTM}(\cdot)$ to shorten Eqn.~(\ref{eq:lstm}) as
{\small\begin{equation}
(\mathbf{h}_{i, t+1}, \mathbf{m}_{i, t+1}) = \text{G-LSTM}(\mathbf{f}_{i,t+1}, \mathbf{h}_{i,t}, \mathbf{m}_{i,t}, \{\mathbf{Z}_{j,t}\}_{j \in \mathcal{N}_{\mathcal{G}}(i)}, \mathbf{W}, \mathbf{U}, \mathcal{G}).
\end{equation} } 
\vspace{-4mm}

\noindent{The} mechanism acts as a memory system, where the information can be written into the memory states and sequentially recorded by each graph node, which is then used to communicate
with the hidden states of subsequent graph nodes and previous Graph LSTM layer. The back propagation is used to train all the weight metrics.

\section{Experiments}

\textbf{Dataset:} We evaluate the performance of the proposed Graph LSTM structure on semantic object parsing on four challenging datasets. 

\textbf{PASCAL-Person-Part dataset~\cite{chen2014detect}.} The public PASCAL-Person-part dataset concentrates on the human part segmentation annotated by Chen et al.~\cite{chen2014detect} from PASCAL VOC 2010 dataset. The dataset contains detailed part annotations for every person. Following~\cite{xia2015zoom}\cite{chen2015attention}, the annotations are merged to be Head, Torso, Upper/Lower Arms and Upper/Lower Legs, resulting in six person part classes and one background class. $1,716$ images are used for training and $1,817$ for testing. The person part of the dataset is particularly challenging because it has large variation in scale and pose. 

\textbf{Horse-Cow parsing dataset~\cite{wang2014semantic}.} The Horse-Cow parsing dataset is a part segmentation benchmark introduced in~\cite{wang2014semantic}. For each class, most observable instances from PASCAL VOC 2010 benchmark~\cite{everingham2012pascal} are manually selected, including 294 training images and 227 testing images. Each image pixel is elaborately labeled as one of the four part classes, including head, leg, tail and body.

\textbf{ATR dataset~\cite{ATR} and Fashionista dataset~\cite{yamaguchi2012parsing}.} Human parsing aims to predict every pixel of each image with 18 labels: face, sunglass, hat, scarf, hair, upper-clothes, left-arm, right-arm, belt, pants, left-leg, right-leg, skirt, left-shoe, right-shoe, bag, dress and null. Originally, 7,700 images are included in the ATR dataset~\cite{ATR}, with 6,000 for training, 1,000 for testing and 700 for validation. 10,000 real-world human pictures are further collected by~\cite{Co-CNN} to cover images with more challenging poses, occlusion and clothes variations. We follow the training and testing settings used in~\cite{Co-CNN}. The Fashionista dataset contains 685 images, among which 229 images are used for testing and the rest for training. 

{\noindent\textbf{Evaluation metric:}} The standard intersection over union (IOU) criterion and pixel-wise accuracy are adopted for evaluation on PASCAL-Person-Part dataset and Horse-Cow parsing dataset, following~\cite{wang2014semantic}\cite{hariharan2014hypercolumns}\cite{chen2015attention}. We use the same evaluation metrics as in~\cite{Yamaguchiparsing13}\cite{ATR}\cite{Co-CNN} for evaluation on two human parsing datasets, including accuracy, average precision, average recall, and average F-1 score.

{\noindent\textbf{Network architecture:}} For fair comparison with~\cite{wang2015joint}\cite{xia2015zoom}\cite{chen2015attention}, our network is based on the publicly available model, ``DeepLab-CRF-LargeFOV"~\cite{chen2014semantic} for the PASCAL-Person-Part and Horse-Cow parsing dataset, which slightly modifies VGG-16 net~\cite{simonyan2014very} to FCN~\cite{long2014fully}. For fair comparing with~\cite{liang2015semantic}\cite{Co-CNN} on two human parsing datasets, the basic ``Co-CNN" structure proposed in~\cite{Co-CNN} is utilized due to its leading accuracy. Our networks based on ``Co-CNN" are trained from the scratch following the same setting in~\cite{Co-CNN}. 

\noindent\textbf{Training: } We use the same data augmentation techniques for the object part segmentation and human parsing as in~\cite{wang2015joint} and~\cite{Co-CNN}, respectively. The scale of the input image is fixed as $321\times321$ for training networks based on ``DeepLab-CRF-LargeFOV". Based on ``Co-CNN", the input image is rescaled to $150\times 100$ as in~\cite{Co-CNN}. We use the SLIC over-segmentation method~\cite{achanta2010slic} to generate averagely 1,000 superpixels for each image. Two training steps are employed to train the networks. First, we train the convolutional layer with $1\times 1$ filters to generate initial confidence maps that are used to produce the starting node and the update sequence for all nodes in Graph LSTM. Then, the whole network is fine-tuned based on the pretrained model to produce final parsing results. In each step, the learning rate of the newly added layers, including Graph LSTM layers and convolutional layers is initialized as 0.001 and that of other previously learned layers, is initialized as 0.0001. All weight matrices used in the Graph LSTM units are randomly initialized from a uniform distribution of [-0.1, 0.1]. The Graph LSTM predicts the hidden and memory states with the same dimension as in the previous convolutional layers. We only use two Graph LSTM layers for all models since only slight improvements are observed by using more Graph LSTM layers, which also consumes more computation resources. The weights of all convolutional layers are initialized with Gaussian distribution with standard deviation as 0.001. We train all the models using stochastic gradient descent with a batch size of 2 images, momentum of 0.9, and weight decay of 0.0005. We fine-tune the networks on ``DeepLab-CRF-LargeFOV" for roughly 60 epochs and it takes about 1 day. For training based on ``Co-CNN" from scratch, it takes about 4-5 days. In the testing stage, one image takes 0.5 second on average except for the superpixel extraction step.

{\noindent\textbf{Reproducibility:}} The proposed Graph LSTM is implemented by extending the Caffe framework~\cite{jia2014caffe}. All networks are trained on a single NVIDIA GeForce GTX TITAN X GPU with 12GB memory. Upon acceptance, we plan to release our source code and trained
models, so that all results  in the paper can be reproduced.

\subsection{Results and Comparisons}

We compare the proposed Graph LSTM structure with several state-of-the-art methods on four public datasets.

\begin{table}[!tp]\setlength{\tabcolsep}{1.2pt}
	\centering\scriptsize
	\caption{Comparison of object parsing performance with four state-of-the-art methods over the PASCAL-Person-Part dataset~\cite{wang2014semantic}. }\label{tab:person}
	\vspace{-1mm}
	\begin{tabular}{cccccccccccccccccccccc}
		\toprule
		{Method} &  head   &  torso  &  u-arms  & l-arms & u-legs & l-legs & Bkg & Avg \\
		\midrule
		DeepLab-LargeFOV~\cite{chen2014semantic}  & 78.09 & 54.02 & 37.29 & 36.85 & 33.73 & 29.61 & 92.85 & 51.78 \\
		HAZN~\cite{xia2015zoom} & {80.79} & {59.11} & {43.05} & {42.76} & 38.99 & 34.46 & 93.59 & 56.11\\
		Attention~\cite{chen2015attention} & {-} & {-} & {-} & {-} & - & - & - & 56.39\\
		LG-LSTM~\cite{liang2015semantic} & \textbf{82.72} & 60.99 & 45.40 & \textbf{47.76} & 42.33 & 37.96 & 88.63 & 57.97\\ 
		\midrule
		\textbf{Graph LSTM} & {82.69} & \textbf{62.68} & \textbf{46.88} & {47.71} & \textbf{45.66} & \textbf{40.93} & \textbf{94.59} & \textbf{60.16} \\
		\bottomrule
		\vspace{-4mm}
	\end{tabular}%
\end{table}%
\begin{table}[!tp]\setlength{\tabcolsep}{2.8pt}
	\centering\scriptsize
	\caption{Comparison of object parsing performance with five state-of-the-art methods over the Horse-Cow object parsing dataset~\cite{wang2014semantic}. }\label{tab:horsecow}
	\vspace{-1mm}
	\begin{tabular}{cccccccccccccccccccccc}
		\toprule
		& & & & \textbf{Horse} & & &\\
		\hline
		{Method} &  Bkg   &  head  &  body  & leg & tail & Fg & IOU & Pix.Acc \\
		\midrule
		SPS~\cite{wang2014semantic}  & 79.14 & 47.64 & 69.74 & 38.85 & - & 68.63 & - & 81.45 \\
		HC~\cite{hariharan2014hypercolumns}  & 85.71 & 57.30 & 77.88 & 51.93 & 37.10 & 78.84 & 61.98 & 87.18 \\
		Joint~\cite{wang2015joint} & 87.34 & 60.02 & 77.52 & 58.35 & {51.88} & 80.70 & 65.02 & 88.49\\
		{LG-LSTM}~\cite{liang2015semantic} & {89.64} & {66.89} & {84.20} & {60.88} & 42.06 & {82.50} & {68.73} & {90.92} \\
		HAZN~\cite{xia2015zoom} & {90.87} & {70.73} & {84.45} & {63.59} & 51.16 & {-} & {72.16} & {-} \\
		\midrule
		\textbf{Graph LSTM} & \textbf{91.73} & \textbf{72.89} & \textbf{86.34} & \textbf{69.04} & \textbf{53.76} & \textbf{87.51} & \textbf{74.75} & \textbf{92.76} \\
		\midrule
		& & & & \textbf{Cow} & & &\\
		\hline
		{Method} &  Bkg   &  head  &  body  & leg & tail & Fg & IOU & Pix.Acc \\
		\midrule
		SPS~\cite{wang2014semantic}  & 78.00 & 40.55 & 61.65 & 36.32 & - & 71.98 & - & 78.97 \\
		HC~\cite{hariharan2014hypercolumns}  & 81.86 & 55.18 & 72.75 & 42.03 & 11.04 & 77.04 & 52.57 & 84.43 \\
		Joint~\cite{wang2015joint} & 85.68 & 58.04 & 76.04 & 51.12 & 15.00 & 82.63 & 57.18 & 87.00\\
		{LG-LSTM}~\cite{liang2015semantic} & {89.71} & {68.43} & {82.47} & {53.93} & {19.41} & {85.41} & {62.79} & {90.43}\\
		HAZN~\cite{xia2015zoom} & {90.66} & \textbf{75.10} & {83.30} & {57.17} & 28.46 & {-} & {66.94} & {-} \\
		\midrule
		\textbf{Graph LSTM} & \textbf{91.54} & {73.88} & \textbf{85.92} & \textbf{63.67} & \textbf{35.22} & \textbf{88.42} & \textbf{70.05} & \textbf{92.43} \\
		\hline
		\vspace{-7mm}
	\end{tabular}%
\end{table}%

\textbf{PASCAL-Person-Part dataset~\cite{wang2014semantic}}: We report the results and the comparisons with four recent state-of-the-art methods~\cite{chen2014semantic}\cite{xia2015zoom}\cite{chen2015attention}\cite{liang2015semantic} in Table~\ref{tab:person}. The results of ``DeepLab-LargeFOV" were originally reported in~\cite{xia2015zoom}. The proposed Graph LSTM structure  substantially outperforms these baselines in terms of average IoU metric. In particular, for the semantic parts with more likely confusions such as upper-arms and lower-arms, the Graph LSTM provides considerably better prediction than  baselines, e.g., $4.95\%$ and $6.67\%$ higher over~\cite{xia2015zoom} for lower-arms and upper-legs, respectively. This superior performance achieved by  Graph LSTM demonstrates the effectiveness of exploiting global context to boost local prediction.

\textbf{Horse-Cow Parsing dataset~\cite{wang2014semantic}}: Table~\ref{tab:horsecow} shows the comparison results with five state-of-the-art methods on the overall metrics. The proposed Graph LSTM gives a huge boost in average IOU. For example, Graph LSTM achieves $70.05\%$, $7.26\%$ better than LG-LSTM~\cite{liang2015semantic} and $3.11\%$ better than HAZN~\cite{xia2015zoom} for the cow class. Large improvement, i.e. $2.59\%$ increase by Graph LSTM in IOU over the best performing state-of-the-art method, can also be observed from the comparisons on horse class. 

\begin{table}[!tp]\setlength{\tabcolsep}{2.8pt}
	\centering\scriptsize
	\caption{Comparison of human parsing performance with seven state-of-the-art methods when evaluating on ATR dataset~\cite{ATR}.}\label{tab:tableoverall}\vspace{-3mm}
	\begin{tabular}{cccccccccccccccccccccc}
		\toprule
		\textbf{Method} &  \textbf{Acc.}   &  \textbf{F.g. acc.}  &  \textbf{Avg. prec.}   &    \textbf{Avg. recall}  &  \textbf{Avg. F-1 score} \\
		\midrule
		Yamaguchi et al.~\cite{yamaguchi2012parsing}& 84.38 & 55.59 & 37.54 & 51.05 & 41.80 \\
		PaperDoll~\cite{Yamaguchiparsing13} & 88.96 & 62.18 & 52.75 & 49.43 & 44.76 \\
		{M-CNN}~\cite{M-CNN} &{89.57} &{73.98} &{64.56} &{65.17} &{62.81}\\
		ATR~\cite{ATR} & {91.11} & {71.04} & {71.69} & {60.25} & {64.38}\\
		{Co-CNN}~\cite{Co-CNN} & 95.23 & 80.90 & 81.55 & 74.42 & 76.95\\
		{Co-CNN (more)}~\cite{Co-CNN} & {96.02} & {83.57} & {84.95} & {77.66} & {80.14}\\
		{LG-LSTM}~\cite{liang2015semantic} & {96.18} & {84.79} & {84.64} & {79.43} & {80.97}\\
		{LG-LSTM (more)}~\cite{liang2015semantic} & {96.85} & {87.35} & {85.94} & {82.79} & {84.12}\\
		CRFasRNN (more)~\cite{crfasrnn} & {96.34} & {85.10} & {84.00} & {80.70} & {82.08}\\
		\midrule
		{Graph LSTM} & {97.60} & {91.42} & {84.74} & {83.28} & {83.76}\\
		\textbf{Graph LSTM (more)} & \textbf{97.99} & \textbf{93.06} & \textbf{88.81} & \textbf{87.80} & \textbf{88.20}\\
		
		\bottomrule
	\end{tabular}%
	\vspace{-2mm}
\end{table}%

\begin{table}[!tp]\setlength{\tabcolsep}{2pt}
	\centering\scriptsize
	\caption{Comparison of human parsing performance with five state-of-the-art methods on the test images of Fashionista~\cite{yamaguchi2012parsing}.}\label{tab:tablefashion}\vspace{-3mm}
	\begin{tabular}{cccccccccccccccccccccc}
		\toprule
		\textbf{Method} &  \textbf{Acc.}   &  \textbf{F.g. acc.}  &  \textbf{Avg. prec.}   &    \textbf{Avg. recall}  &  \textbf{Avg. F-1 score} \\
		\midrule
		Yamaguchi et al.~\cite{yamaguchi2012parsing}  & 87.87 & 58.85 & 51.04 & 48.05 & 42.87 \\
		PaperDoll~\cite{Yamaguchiparsing13}  & 89.98 & 65.66 & 54.87 & 51.16 & 46.80 \\
		ATR~\cite{ATR} & {92.33} & {76.54} & {73.93} & {66.49} & {69.30}\\
		{Co-CNN}~\cite{Co-CNN} & 96.08 & 84.71 & 82.98 & 77.78 & 79.37\\
		{Co-CNN (more)}~\cite{Co-CNN} & {97.06} & {89.15} & {87.83} & {81.73} & {83.78}\\
		{LG-LSTM}~\cite{liang2015semantic} & {96.85} & {87.71} & {87.05} & {82.14} & {83.67}\\
		{LG-LSTM (more)}~\cite{liang2015semantic} & {97.66} & {91.35} & {89.54} & {85.54} & {86.94}\\
		\midrule
		{Graph LSTM} & {97.93} & {92.78} & {88.24} & {87.13} & {87.57}\\
		\textbf{Graph LSTM (more)} & \textbf{98.14} & \textbf{93.75} & \textbf{90.15} & \textbf{89.46} & \textbf{89.75}\\
		
		\bottomrule
	\end{tabular}%
			\vspace{-2mm}
\end{table}%

\textbf{ATR dataset~\cite{ATR}}: Table~\ref{tab:tableoverall} and Table~\ref{tab:F1scores} report the comparison performance with seven state-of-the-arts on overall metrics and F-1 scores of individual semantic labels, respectively. The proposed Graph LSTM can significantly outperform these baselines, particularly, $83.76\%$ vs $76.95\%$ of Co-CNN~\cite{Co-CNN} and $80.97\%$ of LG-LSTM~\cite{liang2015semantic} in terms of average F-1 score. Following~\cite{Co-CNN}, we also take the additional 10,000 images in~\cite{Co-CNN} as extra training images and report the results as ``Graph LSTM (more)". The ``Graph LSTM (more)" can also improve the average F-1 score by $4.08\%$ over ``LG-LSTM (more)". We show the F-1 score for each label in Table~\ref{tab:F1scores}. Generally, our Graph LSTM shows much higher performance than other baselines. In addition, our ``Graph LSTM (more)" significantly outperforms ``CRFasRNN (more)"~\cite{crfasrnn}, verifying the superiority of Graph LSTM over the pair-wise terms in CRF in capturing global context. The results of ``CRFasRNN (more)"~\cite{crfasrnn} are obtained by training the network using their public code.

\textbf{Fashionista dataset~\cite{yamaguchi2012parsing}}: Table~\ref{tab:tablefashion} gives the comparison results on the Fashionista dataset. Following~\cite{ATR}, we only report the performance by training on the same large ATR dataset~\cite{ATR} and then testing on the 229 images of the Fashionista dataset. Our Graph LSTM architecture can substantially outperform the baselines by a large gain. 

\begin{table*}[!tp]\setlength{\tabcolsep}{2.2pt}
	\centering\tiny
	{\caption{Per-Class Comparison of F-1 scores with six state-of-the-art methods on  ATR~\cite{ATR}.}\label{tab:F1scores}\vspace{-3mm}
		
		\begin{tabular}{cccccccccccccccccccccc}
			\toprule
			Method & Hat & Hair & S-gls & U-cloth & Skirt & Pants & Dress & Belt & L-shoe  & R-shoe & Face & L-leg & R-leg & L-arm  & R-arm & Bag & Scarf   \\
			\midrule
			Yamaguchi et al.~\cite{yamaguchi2012parsing} & 8.44 & 59.96 & 12.09 & 56.07 & 17.57 & 55.42 & 40.94 & 14.68 & 38.24 & 38.33 & 72.10 & 58.52 & 57.03 & 45.33 & 46.65 & 24.53 & 11.43\\
			PaperDoll~\cite{Yamaguchiparsing13}& 1.72 & 63.58 & 0.23 & 71.87 & 40.20 & 69.35 & 59.49 & 16.94 & 45.79 & 44.47 & 61.63 & 52.19 & 55.60 & 45.23 & 46.75 & 30.52 & 2.95\\
			{M-CNN}~\cite{M-CNN}  & {80.77} & 65.31 & 35.55& {72.58}& 77.86 &	70.71 &	81.44&	38.45& 	53.87&{48.57}&	72.78&	63.25&{68.24}&{57.40}& 	{51.12}&{57.87}& 43.38\\
			{ATR~\cite{ATR}} & {77.97} & 68.18 & {29.20} & 79.39 & {80.36} & {79.77} & {82.02} & 22.88 & {53.51} & {50.26} & 74.71 & {69.07} & {71.69} & {53.79} & {58.57} & 53.66 & {57.07}\\
			Co-CNN~\cite{Co-CNN} & 72.07 & 86.33 & 72.81 & 85.72 & 70.82 & 83.05 & 69.95 & 37.66 & 76.48 & 76.80 & 89.02 & 85.49 & 85.23 & 84.16 & 84.04 & 81.51 & 44.94\\
			{Co-CNN more}~\cite{Co-CNN} & {75.88} & {89.97} & \textbf{81.26} & {87.38} & {71.94} & {84.89} & {71.03} & 40.14 & {81.43} & {81.49} & {92.73} & {88.77} & {88.48} & {89.00} & {88.71} & {83.81} & {46.24}\\
			{LG-LSTM (more)}~\cite{liang2015semantic}	& {81.13} & \textbf{90.94} & {81.07} & {88.97} & {80.91} & {91.47} & {77.18} & {60.32} & {83.40} & {83.65} & \textbf{93.67} & \textbf{92.27} & {92.41} & {90.20} & \textbf{90.13} & {85.78} & 51.09\\
			\midrule			
			\textbf{Graph LSTM (more)}	& \textbf{85.30} & {90.47} & {72.77} & \textbf{95.11} & \textbf{97.31} & \textbf{96.58} & \textbf{96.43} & \textbf{68.55} & \textbf{85.27} & \textbf{84.35} & {92.70} & {91.13} & \textbf{93.17} & \textbf{91.20} & {81.00} & \textbf{90.83} & \textbf{66.09}\\
			
			\bottomrule
		\end{tabular}
		\vspace{-3mm}
	}
	
\end{table*}

\begin{table}[!tp]\setlength{\tabcolsep}{1.2pt}
	\centering\scriptsize
	\caption{Performance comparisons of using different LSTM structures and taking the superpixel smoothing as the post-processing step when evaluating on PASCAL-Person-Part dataset. }\label{tab:lstm}
	\vspace{-2mm}
	\begin{tabular}{cccccccccccccccccccccc}
		\toprule
		{Method} &  head   &  torso  &  u-arms  & l-arms & u-legs & l-legs & Bkg & Avg \\
		\midrule
		Grid LSTM~\cite{gridlstm} & {81.85} & {58.85} & {43.10} & {46.87} & 40.07 & 34.59 & 85.97 & 55.90\\
		Row LSTM~\cite{van2016pixel}  & 82.60 & 60.13 & 44.29 & 47.22 & 40.83 & 35.51 & 87.07 & 56.80 \\
		Diagonal BiLSTM~\cite{van2016pixel} & 82.67 & 60.64 & 45.02 & 47.59 & 41.95 & 37.32 & 88.16 & 57.62 \\	
		LG-LSTM~\cite{liang2015semantic} & {82.72} & 60.99 & 45.40 & {47.76} & 42.33 & 37.96 & 88.63 & 57.97\\ 
		\midrule
		Diagonal BiLSTM~\cite{van2016pixel} + superpixel smoothing & 82.91 & 61.34 & 46.01 & 48.07 & 42.56 & 37.91 & 89.21 & 58.29 \\	
		LG-LSTM~\cite{liang2015semantic} + superpixel smoothing & \textbf{82.98} & 61.58 & 46.27 & \textbf{48.08} & 42.94 & 38.55 & 89.66 & 58.58\\ 
		\midrule
		\textbf{Graph LSTM} & {82.69} & \textbf{62.68} & \textbf{46.88} & {47.71} & \textbf{45.66} & \textbf{40.93} & \textbf{94.59} & \textbf{60.16} \\
		\bottomrule
		\vspace{-6mm}
	\end{tabular}%
\end{table}%

\vspace{-3mm}
\subsection{Discussions}
\vspace{-1mm}


\textbf{Graph LSTM vs locally fixed factorized LSTM.} Different from the previous locally fixed factorized LSTM structure~\cite{van2016pixel}\cite{liang2015semantic}\cite{gridlstm}, the proposed Graph LSTM adopts adaptive graph topologies for each image and propagates information from different numbers of neighbors to each node. To show the superiority of the Graph LSTM structure more transparently, Table~\ref{tab:lstm} gives the performance comparison among different LSTM structures. These variants use the same network architecture and only replace the Graph LSTM layer with the traditional fixedly factorized LSTM layer, including Row LSTM~\cite{van2016pixel}, Diagonal BiLSTM~\cite{van2016pixel}, LG-LSTM~\cite{liang2015semantic} and Grid LSTM~\cite{gridlstm}. The experimented Grid LSTM~\cite{gridlstm} is a simplified version of Diagnocal BiLSTM~\cite{van2016pixel} where only the top and left pixels are considered. Their  basic structures are presented in Fig.~\ref{fig:graphlstm}. It can be observed that using richer local contexts (i.e., number of neighbors) to update the states of each pixel can lead to better parsing performance. In average, there are six neighboring nodes for each superpixel node in the constructed graph topologies in Graph LSTM. Although the LG-LSTM~\cite{liang2015semantic} has employed eight neighboring pixels to guide local prediction, its performance is still worse than our Graph LSTM. This improvement can be attributed to the adaptive neighborhood topologies and more global context captured by Graph LSTM rather than the number of neighbors.

\textbf{Graph LSTM vs superpixel smoothing.} In Table~\ref{tab:lstm}, we further demonstrate that the performance gain by Graph LSTM is not just from using more accurate boundary information provided by superpixels. The superpixel smoothing can be used as a post-processing step to refine confidence maps by previous LSTMs.  By comparing ``Diagonal BiLSTM~\cite{van2016pixel} + superpixel smoothing" and ``LG-LSTM~\cite{liang2015semantic} + superpixel smoothing" with our ``Graph LSTM", we can find that the Graph LSTM can still bring more performance gain benefiting from its advanced information propagation based on the graph-structured representation.

\begin{table}[!tp]\setlength{\tabcolsep}{1.2pt}
	\centering\scriptsize
	\caption{Performance comparisons with different node updating schemes when evaluating on PASCAL-Person-Part dataset. }\label{tab:traverse}
	\vspace{-2mm}
	\begin{tabular}{cccccccccccccccccccccc}
		\toprule
		{Method} &  head   &  torso  &  u-arms  & l-arms & u-legs & l-legs & Bkg & Avg \\
		\midrule
		BFS (location)  & \textbf{83.00} & 61.63 & 46.18 & \textbf{48.01} & {44.09} & 38.71 & 93.82 & 58.63\\
		BFS (confidence) & {82.97} & {62.20} & {46.70} & {48.00} & 44.02 & 39.00 & 90.86 & 59.11\\
		DFS (location) & 82.85 & 61.25 & 45.89 & 48.02 & 42.50 & 38.10 & 89.04 & 58.23 \\			
		DFS (confidence) & 82.89 & 62.31 & 46.76 & 48.04 & 44.24 & 39.07 & 91.18 & 59.21\\
		\midrule
		\textbf{Graph LSTM (confidence-driven)} & {82.69} & \textbf{62.68} & \textbf{46.88} & {47.71} & \textbf{45.66} & \textbf{40.93} & \textbf{94.59} & \textbf{60.16} \\
		\bottomrule
		\vspace{-3mm}
	\end{tabular}%
\end{table}%

\begin{table}[!tp]\setlength{\tabcolsep}{2pt}
	\centering\scriptsize
	\caption{Performance comparisons of using the confidence-drive scheme based on confidences on different foreground labels when evaluating on PASCAL-Person-Part dataset. }\label{tab:starting}
	\vspace{-1mm}
	\begin{tabular}{cccccccccccccccccccccc}
		\toprule
		{Foreground label} & head  &  torso  &  u-arms  & l-arms & u-legs & l-legs  & Avg\\
		\hline
		Avg IoU & {61.03} & {61.45} & {60.03} & {59.23} & {60.49} & {59.89} & {60.35}\\
		\bottomrule
		\vspace{-4mm}
	\end{tabular}%
\end{table}%

\textbf{Node updating scheme.} Different node updating schemes to update the states of all nodes are further investigated in Table~\ref{tab:traverse}. The Breadth-first search (BFS) and Depth-first search (DFS) are the traditional algorithms to search graph data structures. For one parent node, selecting different children nodes to first update may lead to different updated hidden states for all nodes. Two ways of selecting first children nodes for updating are thus evaluated: ``BFS (location)" and ``DFS (location)" choose the spatially left-most node among all children nodes to update first while ``BFS (confidence)" and ``DFS (confidence)" select the child node with maximal confidence on all foreground classes. We find that using our confidence-driven scheme can achieve better performance than other alternative ones. The possible reason may be that the features of superpixel nodes with higher foreground confidences embed more accurate semantic meanings and thus lead to more reliable global reasoning.   

Note that we use the ranking of confidences on all foreground classes to generate the node updating scheme. In Table~\ref{tab:starting}, we extensively test the performance of using the initial confidence maps of different foreground labels to produce the node updating sequence. In average, only slight performance differences are observed when using the confidences of different foreground labels. In particular, using the confidences of ``head" and ``torso" leads to improved performance over using those of all foreground classes, i.e., $61.03\%$ and $61.45\%$ vs $60.16\%$. It is possible because the segmentation of head/torso are more reliable in the person parsing case, which further verifies that the reliability of nodes in the updating order is important. It is difficult to determine the best semantic label for each task, hence we just use the one over all the foreground labels for simplicity and efficiency in implementation.

\begin{table}[!tp]\setlength{\tabcolsep}{1.2pt}
	\centering\scriptsize
	\caption{Comparisons of parsing performance by the version with or without learning adaptive forget gates for different neighboring nodes when evaluating on PASCAL-Person-Part dataset. }\label{tab:forget}
	\vspace{-1mm}
	\begin{tabular}{cccccccccccccccccccccc}
		\toprule
		{Method} &  head   &  torso  &  u-arms  & l-arms & u-legs & l-legs & Bkg & Avg \\
		\midrule			
		Identical forget gates & \textbf{82.89} & 62.31 & 46.76 & \textbf{48.04} & 44.24 & 39.07 & 91.18 & 59.21\\
		\midrule
		\textbf{Graph LSTM (dynamic forget gates)} & {82.69} & \textbf{62.68} & \textbf{46.88} & {47.71} & \textbf{45.66} & \textbf{40.93} & \textbf{94.59} & \textbf{60.16} \\
		\bottomrule
	\end{tabular}%
		\vspace{-5mm}
\end{table}%

\textbf{Adaptive forget gates.} In the locally fixed factorized LSTMs, the same forget states are learned to  exploit the influences of neighboring pixels on the updated states of each pixel. Whereas in Graph LSTM, adaptive forget gates are adopted to treat the local contexts from different neighbors differently. The superiority of using adaptive forget gates can be verified in Table~\ref{tab:forget}. ``Identical forget gates" shows the results of learning identical forget gates for all neighbors and simultaneously ignoring the memory states of neighboring nodes. Thus in ``Identical forget gates", the $g_i^f$ and $\mathbf{m}_{i,t+1}$ in Eqn.~(\ref{eq:lstm}) can be simply computed as

\vspace{-5mm}
\begin{equation}
\begin{split}
g^f_{i} = & \delta(W^f\mathbf{f}_{i,t+1} + U^f\mathbf{h}_{i,t} + U^{fn}\bar{\mathbf{h}}_{i,t} + b^f),\\
\mathbf{m}_{i, t+1} =& g^f_{i}\odot \mathbf{\mathbf{m}}_{i,t} +  g^u_i \odot g^c_i.\\
\end{split}
\label{eq:identical}
\end{equation}
\vspace{-3mm}

It can be observed that learning adaptive forgets gates in Graph LSTM shows better  performance over learning identical forget gates for all neighbors on the object parsing task, as diverse semantic correlations with local context can be considered and treated differently during the node updating. Compared to Eqn.~(\ref{eq:identical}), no extra parameters is brought to specify adaptive forget gates due to the usage of the shared parameters $U^{fn}$ in Eqn.~(\ref{eq:lstm}).

\begin{figure}[!tp]
	\begin{center}
		\includegraphics[scale=0.48]{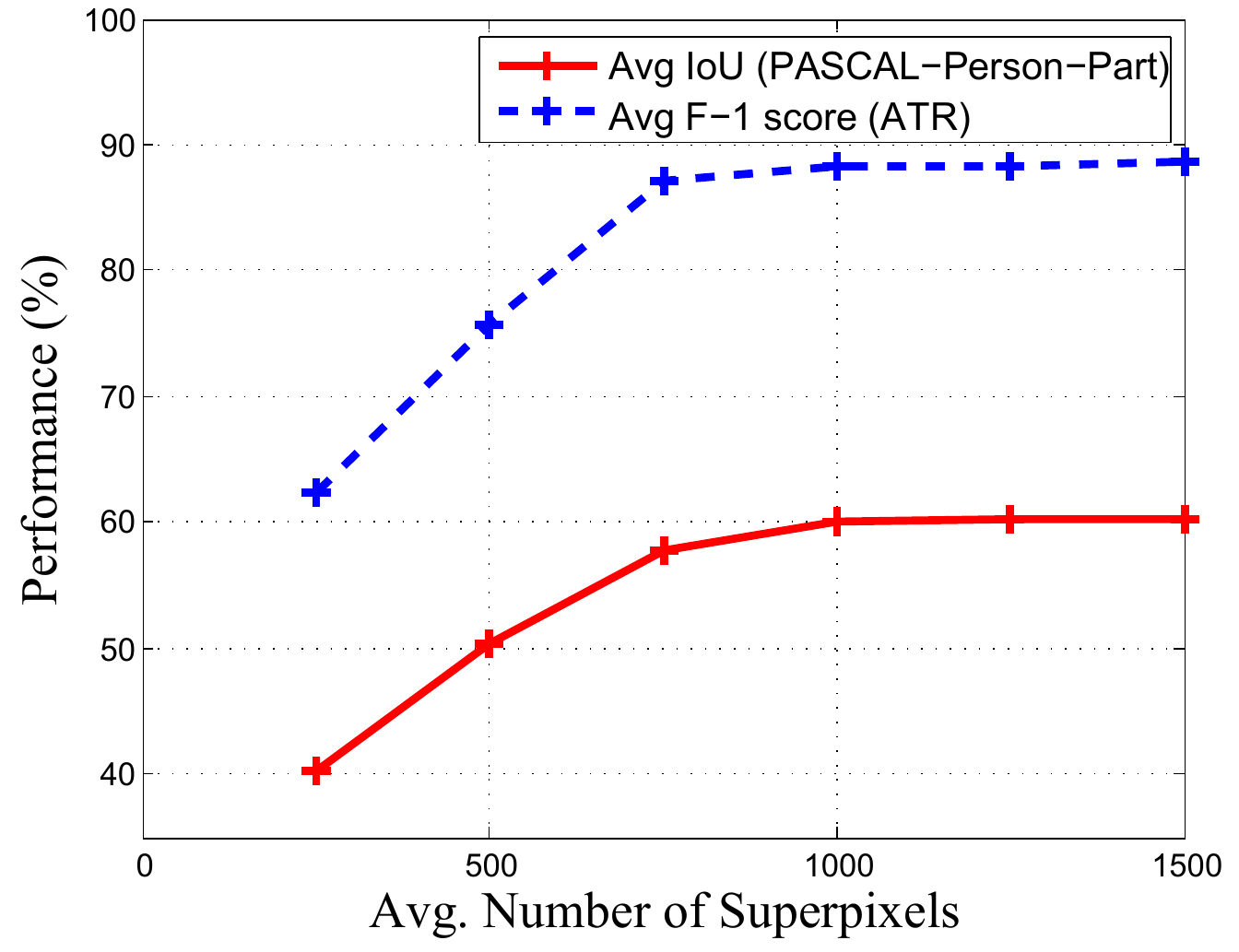}\vspace{-4mm}
		\caption{{Performance comparisons with six averaged numbers of superpixels when evaluating on PASCAL-Person-Part and ATR datasets, including 250, 500, 750, 1000, 1250, 1500 . }}
		\label{fig:super}
	\end{center}
	\vspace{-6mm}
\end{figure}

\textbf{Superpixel number.} The superpixels can pre-group pixels based on spatial and appearance similarity, which reduce the number of elements and keep semantic consistency. However, the drawback is that superpixels may introduce quantization errors whenever pixels within one superpixel have different ground truth labels. We thus evaluate the performance of using different average numbers of superpixels to construct the graph structure. As shown in Fig.~\ref{fig:super}, there are slight improvements when using over 1,000 superpixels. We thus use averagely 1,000 superpixels for each image in all our experiments by balancing the computation efficiency and accuracy.

\textbf{Residual connections.} Residual connections were first proposed in~\cite{he2015deep} to better train very deep convolutional layers. The version in which the residual connections  are eliminated achieves $59.12\%$ in terms of Avg IoU on PASCAL-Person-Part dataset. It demonstrates that residual connections between Graph LSTM layers can also help boost the performance, i.e., $60.16\%$ vs $59.12\%$. Note that our Graph LSTM version without using residual connections is still significantly better than all baselines in Table~\ref{tab:person}.

\vspace{-1mm}
\subsection{More Visual Comparison and Failure cases}
\vspace{-1mm}

\begin{figure}[!tp]
	\begin{center}
		\includegraphics[scale=0.55]{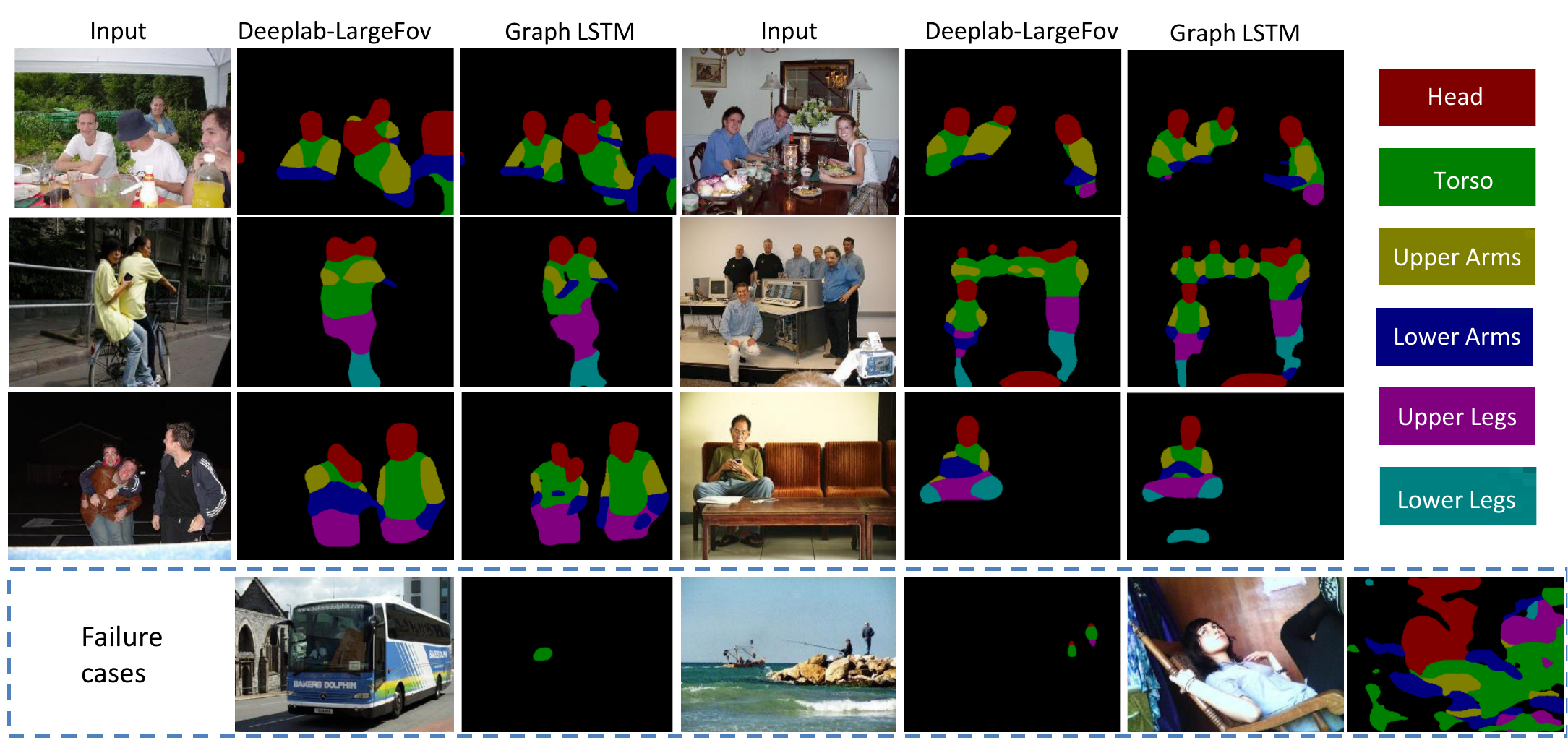}\vspace{-4mm}
		\caption{{Comparison of parsing results of our Graph LSTM and the baseline ``DeepLab-LargeFov" and some failure cases by our Graph LSTM on PASCAL-Person-Part.}}
		\label{fig:person}
	\end{center}
	\vspace{-6mm}
\end{figure}

\begin{figure}[!tp]
	\begin{center}
		\includegraphics[scale=0.75]{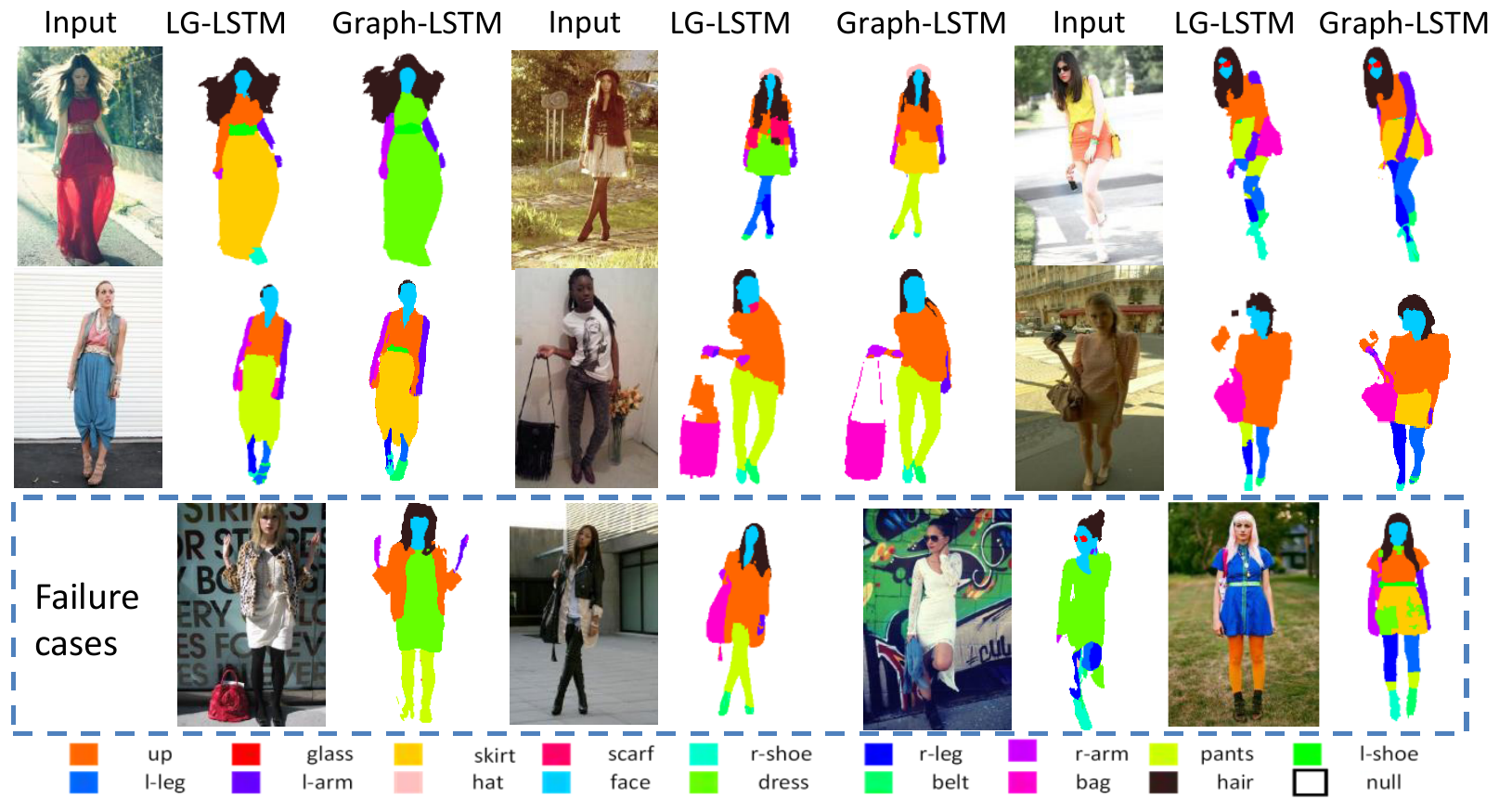}\vspace{-4mm}
		\caption{{Comparison of parsing results of our Graph LSTM and the LG-LSTM~\cite{liang2015semantic} and some failure cases by our Graph LSTM on ATR dataset.}}
		\label{fig:ATRresults}
	\end{center}
	\vspace{-8mm}
\end{figure}

The qualitative comparisons of parsing results on PASCAL-Person-Part and ATR dataset are visualized in Fig.~\ref{fig:person} and Fig.~\ref{fig:ATRresults}, respectively. In general, our Graph-LSTM outputs more reasonable results for confusing labels by effectively exploiting global context to assist the local prediction. We also show some failure cases on each dataset, and find that our Graph LSTM has difficulty in segmenting semantic parts for very small objects (as shown in Fig.~\ref{fig:person}) and parts with very similar appearances (e.g., the shoes and pants in the second failure image in Fig.~\ref{fig:ATRresults}.).

\vspace{-1mm}
\section{Conclusion and Future Work}
\vspace{-1mm}

In this work, we proposed a novel Graph LSTM network to address the fundamental semantic object parsing task. Our Graph LSTM generalizes the existing LSTMs into the graph-structured data. The adaptive graph topology for each image is constructed by connecting the arbitrary-shaped superpixels nodes via their spatial neighborhood connections. The confidence-driven scheme is used to adaptively select the starting node and determine the node updating sequence. The Graph LSTM can thus sequentially update the states of all nodes. Comprehensive evaluations on four public semantic object parsing datasets well demonstrate the significant superiority of our graph LSTM. In the future, we will explore how to dynamically adjust the graph structure to directly produce the semantic masks according to the connected superpixel nodes.

\clearpage

\bibliographystyle{splncs}
\bibliography{egbib}
\end{document}